# Assessing thermal imagery integration into object detection methods on ground-based and air-based collection platforms

**JAMES E GALLAGHER[1], EDWARD J OUGHTON**
[1] Department of Geography & Geoinformation Science Department, George Mason University, Fairfax, VA 22030, USA, e-mail: (jgalla5@gmu.edu)
[2] Department of Geography & Geoinformation Science Department, George Mason University, Fairfax, VA 22030, USA, e-mail: (eoughton@gmu.edu)

Corresponding author: James E. Gallagher (e-mail: jgalla5@gmu.edu).

This work was supported by the Geography & Geoinformation Science Department at George Mason University.

**ABSTRACT** Object detection models commonly deployed on uncrewed aerial systems (UAS) focus on identifying objects in the visible spectrum using Red-Green-Blue (RGB) imagery. However, there is growing interest in fusing RGB with thermal long wave infrared (LWIR) images to increase the performance of object detection machine learning (ML) models. Currently LWIR ML models have received less research attention, especially for both ground- and air-based platforms, leading to a lack of baseline performance metrics evaluating LWIR, RGB and LWIR-RGB fused object detection models. Therefore, this research contributes such quantitative metrics to the literature .The results found that the ground-based blended RGB-LWIR model exhibited superior performance compared to the RGB or LWIR approaches, achieving a mAP of 98.4%. Additionally, the blended RGB-LWIR model was also the only object detection model to work in both day and night conditions, providing superior operational capabilities. This research additionally contributes a novel labelled training dataset of 12,600 images for RGB, LWIR, and RGB-LWIR fused imagery, collected from ground-based and air-based platforms, enabling further multispectral machine-driven object detection research.

## I. INTRODUCTION

Despite the recent growth and proliferation of machine learning (ML) object detection algorithms, most approaches commonly focus on the visible light portion of the electromagnetic spectrum, for example, using Red-Green-Blue (RGB) images [1] [2] [3] [4] Hitherto, long wave infrared (LWIR) spectrum has received less research attention for ML object detection activities. While machine-assisted RGB models are effective during daytime periods, machine-assisted LWIR-models are generally more effective at night or during periods of decreased visibility [5] [6] [7] [8]. Given the contrasting strengths and weaknesses between RGB and LWIR, a growing area of research examines the blending of these different capabilities with the ultimate aim of providing superior object detection techniques [9] [10] [11]. Most software techniques for developing both RGB and LWIR models are similar. However, the accessibility and cost of LWIR sensors has made it difficult to commonly implement multispectral approaches, limiting research activities.

Another technology that is rapidly proliferating and becoming easier to access is commercially available off-the-shelf drones [12]. For example, in industry, government and military applications Uncrewed Aerial Systems (UAS) are having a profound impact on quantity and breadth of imagery collection [13]. Increasingly, UASs are being outfitted with not only RGB sensors to collect overhead imagery, but also with an array of other infrared related sensors to collect valuable multispectral data [14].

Given these limitations, the literature currently lacks scientific evaluation metrics on how different image fusion techniques affect model performance when performing object detection from drone platforms. Therefore, this research intends to investigate the following research question:

1. How do fused RGB-LWIR object detection models perform against separate RGB and LWIR approaches?

2. What is the impact of image processing (IP) on RGB, LWIR and fused RGB-LWIR model performance?

3. In what way do different drone altitudes affect model performance?

The aim of this research project is therefore to develop the fundamental scientific research which supports the design of automated machine-driven object detection when integrating thermal imagery (long wave infrared), focusing initially on identifying civilian objects classes (humans, cars, trucks) as a proof of concept for future industrial, government and military applications.

The contribution of this research is to provide new quantitative scientific information for how RGB, RGB-LWIR and LWIR object detection models perform via various drone airframes (specifically, twelve different types of mixed image models, collected through a variety of ground-based air-based platforms).

This is important because the use of both multispectral object detection models and UAS continue to grow at rapid rates, and the findings of this research support future use case decisions. Indeed, understanding the strengths and weaknesses of how certain ML models perform on various airframes is necessary for effective implementation of these two technologies.

In the following section a literature review is undertaken, followed by the articulation of a method capable of answering the research questions in Section III. Next, in Section IV the results are presented, before returning in the discussion to reevaluate the research questions in Section V. Finally, conclusions are provided in Section VI.

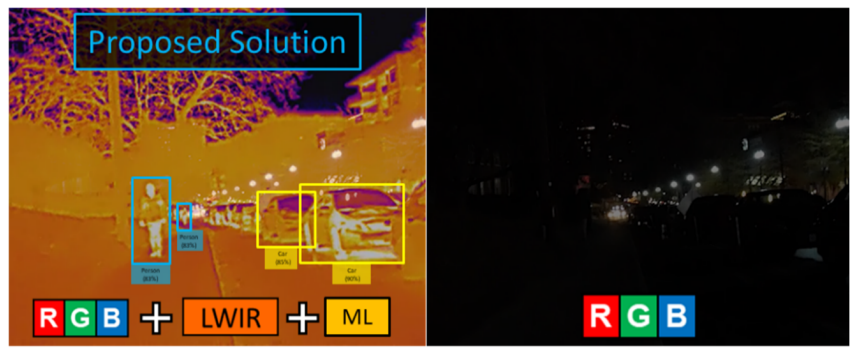

**FIGURE 1.** Example of the proposed solution on the left (LWIR) when compared to an associated RGB image.

## II. LITERATURE REVIEW

The existing literature identifies two key benefits for integrating LWIR with RGB to enhanced ML object detection models. Firstly, RGB sensors are limited in their capacity to detect in low visibility settings, or in situations where visibility is limited due to foliage, smoke or fog [15] [16]. Therefore, integrating LWIR imagery enhances both human and machine three-dimensional (3D) depth perception when compared to traditional RGB imagery, providing an overall increase in situational awareness.

Secondly, LWIR sensors are superior at segmenting the object of interest from the image background (edge detection) [16], provided that the object of interest is radiating a thermal signature. LWIR object detection is primarily discussed in military and homeland security use cases to detect illicit activity and identify targets, especially at night [17], [18]. As previously mentioned, most IR sensors for military and national security applications use NIR, which does not work well for ML object detection models.

In terms of the wider literature, one recent study evaluated ML object detection models that analyzed RGB and LWIR imagery to better identify humans from a ground-based system [17]. When attempting to identify humans in adverse weather conditions the LWIR model achieved a mAP of 97.9% while the RGB model achieved a mAP of 19.6% [17]. Indeed, both LWIR and RGB models were tested, although no baseline performance metrics were provided for a blended RGB-LWIR approach. The research used ground-based sensors and utilized YOLOv3, which stands for 'you only look once-version 3,' and a thermal dataset to attempt to identify humans and animals during various weather conditions that include clear, foggy and rainy days. Although their LWIR model outperformed the RGB model, the performance gap was most significant when visibility was limited. The thermal ML model was also highly accurate in differentiating multiple object classes in a single image, reaching a recall of 98% with a F1 score of 97% [17].

A separate research study recently used LWIR imagery to train an object detection model that achieved an average accuracy of 91.9% during periods of limited visibility [19]. However, it was identified that a shortfall of LWIR object detection is that LWIR cameras have issues identifying object classes at distance. As the object class is farther away, the thermal edges begin to blur and the thermal signature resolution deteriorates, making it difficult for the ML model to conduct edge detection [20]. Thus, because of this resolution decrease over distance, this supports the conjecture that fusing RGB with LWIR provides additional value in model performance.

Another benefit to RGB-LWIR fusion is the ability to adjust fusion levels between the RGB-LWIR sensors as ambient and ground temperatures increase, creating an effect called thermal crossover. When the target object is the same temperature as the ground, thermal cross over takes place leading to a loss of contrast between the target object and the ground [21]. Depending on the environment and season, thermal crossover typically occurs twice a day. Via a ground based LWIR ML object detection model approach, thermal crossover is not as large an issue because the horizon provides a dark background to contrast against thermal target objects. However, from a UAS the bird's-eye view of the ground offers significantly lower contrast with the target object. When using a LWIR camera without an RGB camera or having the ability to conduct RGB-LWIR fusion, the ambient and ground temperature must be factored in prior to flight.

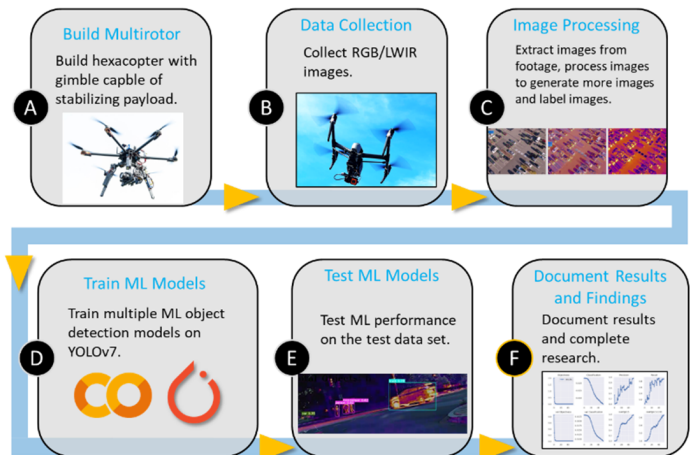

**FIGURE 2.** Workflow for the research method visualized.

Another research study that used LWIR sensors from a low-flying multirotor quadcopter collected thermal data to create a human detection model that identifies human heat signatures. The approach was applied to a rescue operations use case following natural disasters by using object segmentation and fusion technique called 4-channel [22]. The 4-channel ML model conducted "early fusion" of RGB-thermal images, performing better than the traditional "late fusion" model. This study focused on object segmentation of LWIR images taken from the UAS post-flight and did not conduct object detection from LWIR images or RGB-LWIR fused images.

The reliability of LWIR sensors to work in complex environments has led to adoption in numerous technologies. For example, LWIR sensors are used to advance semantic segmentation, classifying pixels in an image associated to a label class, with key use cases in autonomous driving [23] [24], [25]. However, a key issue in the application of this technology to autonomous driving is the low resolution and heavy noise present in LWIR images, when compared to RGB methods [26].

Thermal object detection is also advantageous because of the ability to conform an image to a desired color palette [27], thereby reducing the overall number of colors compared to RGB images [28]. Often, RGB images can have backgrounds that blend in with the object of interest [29], making object detection a more challenging task. In contrast, thermal imagery highlights the object of interest and provides consistent color returns despite background variations and changing visibility conditions [30].

RBG-LWIR fusion is also discussed in semantic segmentation and salient object detection [31], [32]. Semantic segmentation is highly effective at parsing a raw image and ascertaining pixel values to accurately identify multiple object classes in an image [33]. RGB-LWIR fusion is also highly beneficial for computer vision because the fusion of these two sensors provides the rich color, texture and context from RGB images while simultaneously illuminating the edges of the object class derived from LWIR images [34], [35]. There are also studies that discuss improving the RGB-LWIR cross-modal fusion process to enhance scene parsing through more defined edge-detection [36], [37]. To date, existing literature only discusses RGB-LWIR fusion in equal parts (50/50). An RGB-LWIR fusion approach has also been commonly applied to undertake 3D mapping of building interiors and exteriors, to quantify key built environment aspects such as energy efficiency [38].

LWIR based object detection does present several key challenges for ML algorithms. One such issue is blurring in LWIR imagery caused by object movement or LWIR camera movement [39]. One study addressed this issue using a LWIR image restoration algorithm that conducts super-resolution reconstruction and deblurring while simultaneously running the object detection algorithm [39]. Although the methods to deblur LWIR images does increase the overall accuracy of the object detection results, it also requires increased computer processing to conduct simultaneous image restoration and object detection when conducting real-time inference on edge devices. In this research study there is an undetermined level of image blurring induced by the moving airframe with RGB-LWIR cameras.

Another issue with LWIR object detection is that there exists a shortage of publicly available LWIR datasets or pre-trained LWIR models [40]. Indeed, there are multiple pre-trained RGB ML models and datasets to choose from, but very few LWIR datasets and pre-trained models. Labelled LWIR datasets are scarce because they are expensive to collect and produce, and LWIR cameras are not widely available to the same degree as RGB cameras [40] [41]. One idea proposed in the literature to help grow LWIR datasets is to have an RGB camera aligned with a LWIR camera to rapidly collect and automatically label objects of interest using transfer learning [42]. Although this automated collection and labeling research may only be conducted from ground-based sensors, such an approach could also be used from aerial platforms to simultaneously collect data and train RGB and LWIR ML models, thus creating a pre-trained RGB-LWIR ML model for UAS applications. Another method to grow a LWIR imagery dataset is to create artificial LWIR images by applying IP techniques to RGB images to make them appear as LWIR images. Although this is not a perfect solution, this method is able to produce very large synthetic LWIR datasets in a cost efficient way [43].

An IP technique that is mentioned in the existing literature is fusing LWIR imagery with RGB imagery both in hardware and in software to increase object detection accuracy [44]. One such study deployed a model trained on fused RGB-LWIR imagery and used a camera capable of conducting RGB-LWIR fusion to count potholes and determine pavement conditions. The fused RGB-LWIR ML model was able to achieve a surprisingly high precision score of 98.3% [44]. RGB-LWIR fusion is also used in ML models to create resilient object detection models that can continue to detect objects when visibility deteriorates [5]. Although LWIR imagery does deteriorate slightly due to atmospheric turbulence (wind, velocity, humidity, etc.) and distance, the image deterioration for LWIR images is still significantly less when compared to RGB image deterioration [45].

There are also examples of RGB-LWIR fused object detection models being successfully used during the COVID-19 pandemic to automatically identify humans with abnormally high temperatures [46]. RGB-LWIR fusion has also been tested by wildlife and forest management to help more accurately detect tree-trunk diameters from ground-based uncrewed vehicles [47]. There is some discussion in the literature on the amount of RGB-LWIR image fusion that is optimal for object detection and object tracking. For example, an RGB image can be dominant over a LWIR image (low fusion), or both RGB and LWIR images can be equally fused together (equal fusion) [48]. Poor RGB-LWIR fusion models can sometimes lead to contaminated images. For example, where the object edges from RGB images do not line up with the edges of the companion LWIR image, leading to poor performing ML models [49]. This is due mostly to parallax, which is image distortion due to two different fields of view (FoV) [50]. High-end thermal imagers have duel RGB-LWIR cameras on them, allowing for image fusion in hardware [51]. These dual-sensor cameras also suffer from a degree of parallax. Although using a dual sensor RGB-LWIR camera would be optimal for this research study and for real-time RGB-LWIR fused object detection, it was not utilized due to the high cost of these dual sensor RGB-LWIR cameras.

In terms of object detection ML models on UAS, there is a plethora of literature that discusses implementing RGB object detection and object tracking models on UAS [52], but no discussion of performance metrics comparing LWIR object detection versus RGB object detection models on a UAS. Conducting object detection from a UAS, whether it be RGB or LWIR object detection, is inherently a difficult task for computer vision because objects are seen in various scales and backgrounds [53]. Despite this challenge, object detection and object tracking ML models on UAS continue to advance [54]. Some UAS object detection models even possess the capability to conduct multiple object tracking [55] as well as provide relevant metadata such as object speed, trajectory and coordinates that are within 5-15 m accuracy [56]. Lastly, to reiterate, no research to date has been conducted on measuring RGB, LWIR, and RGB-LWIR fused object detection models from UAS video footage.

## A. APPLICATIONS

There is a plethora of use-cases for how fused RGB-LWIR ML models on a UAS can be leveraged to solve spatial problems across diverse disciplines. For example, companies and industries lack the capability to conduct wide-area ML powered thermal object detection [57]. The oil industry lacks the capability to conduct long range thermal classification of its pipelines to measure pipeline integrity and health [58], in order to minimize potential infrastructure disruption [59] [60]. Electric companies lack the capability to conduct drone-based real-time ML thermal surveys of expansive electrical grids and electrical substations [61]. Search and rescue teams lack a deployable UAS with onboard ML object detection that can conduct large area scans to find the thermal signatures of victims [62].

Large-scale thermal object detection from UAS can also be highly beneficial to the agriculture sector for tracking livestock and large herds, as well as provide early warning of the presence of animals that can cause damage to fields and livestock [63]. Military and law enforcement entities will also benefit from the results of this study to improve their collection and surveillance activities, especially collection activities involving footage derived from UAS and closed-circuit television [64], [65]. Aside from increasing detection accuracy with RGB-LWIR fused object detection, ML models on UAS can process data in real time [66], providing decision makers with actionable information, as well as preventing the need for a team of imagery analysts to spend hours processing overhead imagery and video data [67].

## B. NIR VERSUS LWIR

Most UAS used by industries, governments and militaries generally lack a robust variety of onboard sensors. When discussing infrared payloads on sUAS, the most prevailing infrared sensor is near infrared (NIR) [68]. NIR is almost exclusively used for night flying and provides a grayscale image. [69] [70].

The NIR grayscale image makes it both difficult to detect objects and conduct edge detection [71]. The latter is particularly critical for developing automated object detection via machine learning models [72]. To conduct object detection with ML, especially when collecting during periods of darkness, a LWIR camera needs to be onboard the UAS [73]. It is worth noting that although LWIR cameras work best at night or early mornings because of cooler ambient temperatures surrounding the thermal target, LWIR cameras still work during the day and can be fused with RGB images to enhance edge detection while retaining important details from the RGB image [74]. Next, a method is presented which is capable of answering the research questions posed for this research.

## III. METHOD

This method describes six key steps (build multirotor, collect data, image processing, train models, test models, and evaluate findings) which when combined produce a final set of model performance metrics.

### A. BUILD MULTIROTOR

A multirotor hexacopter UAS is built using custom components, 3D printed parts and open-source code. For the hexacopter, the Tarot T960, which is a 6-arm collapsible frame with six 45.7 cm (18 in) propellers and has an approximate endurance time of twenty minutes [75], is used to collect overhead data. This frame-propeller combination is capable of carrying and stabilizing the RGB-LWIR payload. A 3-axis gimbal will also be used to stabilize the sensors in flight to help reduce image distortion from aircraft vibrations [76]. The hexacopter will also be fitted with a Pixhawk Cube Orange as the flight controller to ensure altitude and speed variables are being accurately controlled [77]. When collecting data near humans, a DJI Inspire 2 is used. The DJI Inspire 2 is safer to operate around humans but does not have the 3-axis gimbal to stabilize the RGB-LWIR payload. The RGB-LWIR payload is attached to the bottom of the DJI Inspire 2 with hook-and-loop fasteners and the camera angle is adjusted manually after each flight.

### B. DATA COLLECTION

For ground-based data collection the RGB-LWIR sensors are mounted onto a bike, allowing for higher mobility and collecting data from various scenes, thus increasing data diversity to help build a resilient ground-based ML model. Data is collected during various times of the day at different temperatures to ensure data diversity. Atmospheric metrics are measured at the time of data collection to determine wind speed (anemometer) and illumination (illuminance meter). This metadata will help quantify model performance in atmospheric conditions. Although night data was collected, no night-data was used in the training process or in the test images.

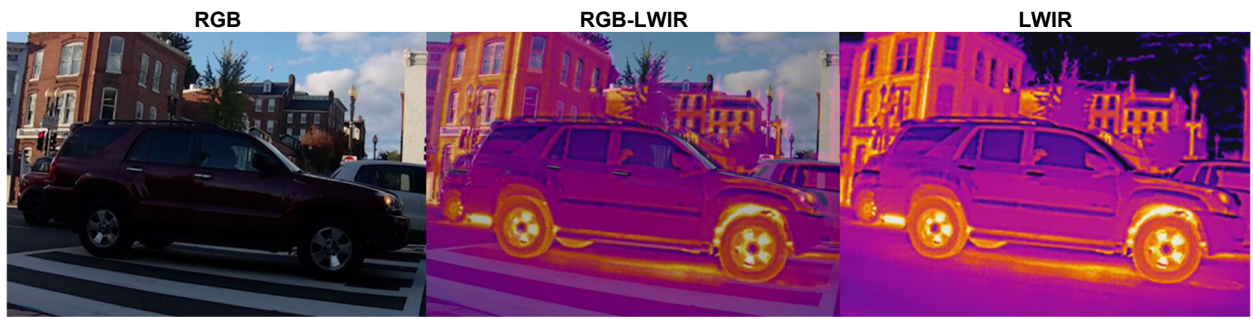

**FIGURE 3.** An example of an RGB image (left), an RGB-LWIR fused image (middle) and a LWIR image (right).

Overhead imagery collection for the air-based ML models is collected from the multirotor hexacopter and DJI Inspire 2 (figure 6). The hexacopter offers the best platform in terms of stability and control to collect RGB and LWIR imagery. The RGB and LWIR camera on the multirotor is co-aligned and maintain the same field of view to reduce parallax and to ensure that similar images are being collected between the two sensors [78]. Both sensors are recording in full-motion video at 30 frames per second (FPS) during flight. Footage is recorded and extracted on the camera's micro-SD cards. Frames of interest from the footage will then be extracted and converted into images to train the ML model. Images will also be collected from various altitudes (15 m, 30 m, 45 m, 61 m, 76 m, 91m, 106m, 122m) to ensure image diversity and to help reduce model performance loss at higher altitudes. A thermal palette called fusion was selected for this research. The fusion palette highlights the most radiant objects in a frame as orange and the least-radiant objects as purple. This palette provides the most contrast when fused with RGB images.

**Image Processing (IP) Techniques**

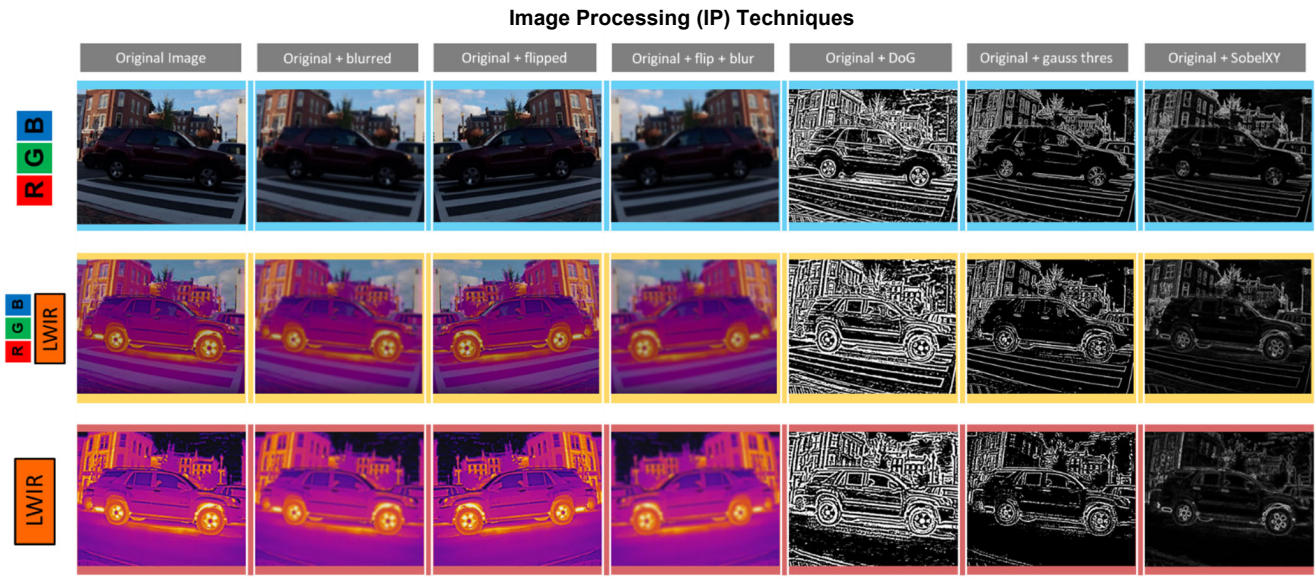

**FIGURE 4.** This chart illustrates the six image processing techniques used to enhance the ML models. The DoG, Gaussian threshold and Sobel-XY image processing techniques helped to amplify the edges of the object class, thus leading to better edge detection and higher mAP.

The images are collected from various camera angles at different times of the day, thus creating a more resilient RGB-LWIR model [79]. For data collection, 100 RGB images and 100 LWIR images are extracted from the full-motion video footage with each object class, (i) car, (ii) truck, (iii) person. The RGB and LWIR footage is then fused in Adobe Premier Pro with a 50-50 fusion ratio to create an additional 100 images for the fused RGB-LWIR dataset. The RGB resolution is also cut in half from 1280x1024 pixels to 640x512 pixels to conform the RGB resolution to the LWIR resolution.

Video footage data will also be collected to evaluate model performance. Data from this video footage will not be used to train the models. Rather, this video footage will be used to measure inference levels in various environments and altitudes to determine model behavior and characteristics. Ground-based model inference will be evaluated against footage from traffic in Dupont Circle, Washington D.C., in the month of August 2022. Air-based model inference performance will be evaluated on video footage that is collected between November and December of 2022 in Manassas, Virginia, near a major highway (Interstate 66). To evaluate inference performance at various altitudes video footage will be collected in December of 2022 in Fairfax, Virginia, at a hardware store parking lot where all object classes are present.

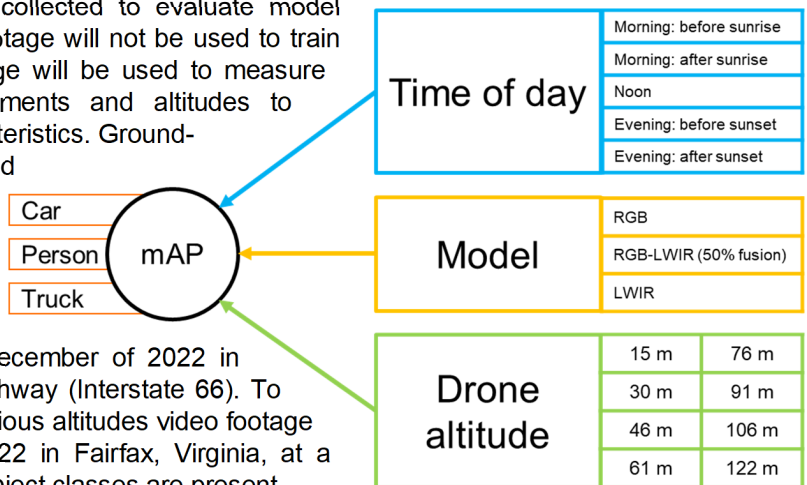

**FIGURE 5.** The research model

### C. IMAGE PROCESSING, AND LABELING

IP techniques are then applied to the original images to increase the number of images in the training dataset while simultaneously generating edge-enhanced images to increase model performance [80]. Six image processing and mathematical morphology techniques are carried out on the original images to both increase the training dataset and enhance model performance. The IP techniques used are flipping, blurring, blurring & flipping, Gaussian thresholding, difference of Gaussians (DoG) and Sobel-XY (figure 4). These IP techniques will help increase model performance on both ground and air platforms. The blurred and blurred + flipped IP techniques are especially useful because of video vibrations caused by the oscillatory motions from the aircraft's propellers [81]. Model training on blurred images helps to ensure that the model will continue to work when frames are blurred due to camera movement, target object movement, or both. Although counterintuitive, training ML models with blurred images tend to increase detection rates and confidence levels [82]. All processing code for generating and exporting augmented images is collated in a Google Colab notebook for reproducibility [83].

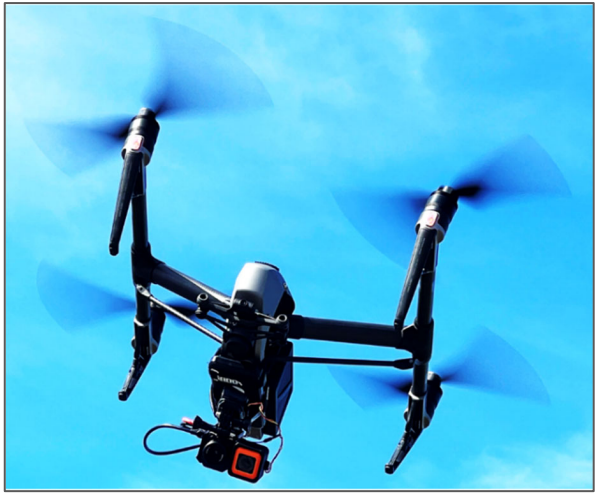

**FIGURE 6.** The primary air-based platform used for this research, the DJI Inspire 2 carrying the RGB-LWIR payload.

After image processing, a total of 10,800 new images are generated, resulting in a total of 12,600 total images. 90% of the dataset (11,340 images) are used for training, 5% (630 images) are used for validation. The remaining 5% (630 images) are used for testing. None of the newly generated images except original+flipped images are used for testing. This is to ensure that testing results are similar across all ML models. As mentioned previously, the imagery collected between the LWIR and RGB cameras will need to be conformed to the same resolutions to conduct ML training on both datasets. The thermal camera, the FLIR Vue Pro R, has a resolution of 640 x 512 pixels, while the RGB camera, the RunCam 5 Orange, has a resolution of 1280 x 1024 pixels. Resolution is adjusted in Adobe Premier pro. Lastly, all images are labeled using LabelImg, which is an open-source python based image labeler [84].

### D. BUILDING THE OBJECT DETECTION MODEL

This research will utilize YOLOv7 as the Convolutional Neural Network (CNN) to perform object detection. YOLOv7 is built with PyTorch and uses the COCO dataset as the backbone [85]. YOLOv7 was selected because to date it surpasses all existing object detectors in terms of speed and accuracy [86]. YOLOv7 was released by Wong Kin Yiu in 2020 and is considered one the fastest open-source object-detection models at the time this research was conducted [87], [88]. The YOLOv7-E6 object detection model outperforms transformer-based SWIN-L Cascade-Mask r-CNN with a 509% increase in speed and 2% increase accuracy, as well as outperforms YOLOR, YOLOX, scaled-YOLOv4, YOLOv5 and many other object detectors [86].

A primary shortfall of the YOLO family of object detection is that YOLO struggles to detect smaller objects within an image, which is primarily due to spatial constraints in the algorithm [89]. There are seven variants of YOLOv7. However only three variants are capable of being deployed on edge devices. The three edge-deployable variants are YOLO7-Tiny, YOLOv7 and YOLOv7-W6 [90]. The standard YOLOv7 variant is used for this research study [91].

Roboflow and Pytorch are also implemented when training the custom object detection model. The model is trained through 55 epochs. This number was selected to prevent overtraining. There is an imbalance in the number of car, truck and people labels in the dataset, making overfitting a possibility if the models are trained through too many epochs [92]. Cars have the most labels in the dataset while trucks have the least. Training the dataset through more than 55 epochs may result in an increase in car false positives, thus decreasing the mean average precision (mAP) of the model.

Twelve ML (table 1) models are created using (i) ground-based IP-LWIR images, (ii) ground-based IP-RGB images, (iii) ground-based IP-RGB-LWIR, (iv) ground-based LWIR images, (v) ground-based RGB images, (vi) ground-based RGB-LWIR images, (vii) air-based IP-LWIR images, (viii) air-based IP-RGB images, (ix) air-based IP-RGB-LWIR images, (x) air-based LWIR images, (xi) air-based RGB images, and (xii) air-based RGB-LWIR images. Each model uses images that share the same vantage point and similar IP techniques to ensure consistency, controlling for key experimental uncertainties. Hitherto, all models will be referenced by the model names in table 1.

| Platform | Model Type | IP | Model name |
|---|---|---|---|
| Ground (G) | RGB | Yes (IP) | G_RGB_IP |
| | RGB | No | G_RGB |
| | RGB + LWIR | Yes (IP) | G_RGB-LWIR_IP |
| | RGB + LWIR | No | G_RGB-LWIR |
| | LWIR | Yes (IP) | G_LWIR_IP |
| | LWIR | No | G_LWIR |
| Air (A) | RGB | Yes (IP) | A_RGB_IP |
| | RGB | No | A_RGB |
| | RGB + LWIR | Yes (IP) | A_RGB-LWIR_IP |
| | RGB + LWIR | No | A_RGB-LWIR |
| | LWIR | Yes (IP) | A_LWIR_IP |
| | LWIR | No | A_LWIR |

TABLE 1. Model types with model name abbreviations

## IV. RESULTS

The first results generated were based on how the models performed against the 5% non-processed test image batch, that was part of ML training process. Results were automatically generated upon completion of the training process. Each ML object detection model took approximately two hours two train on a Google Colaboratory Pro+ account with premium GPU access. The mean average precision (mAP) is the primary metric analyzed following testing, which is a popular approach for comparing ML object detection performance between detection algorithms, such as YOLO, R-CNN and MobileNet SSD etc. [93]. mAP measures localization (location of the object class within an image) and classification (identifying the object being classified). When both air and ground mAP results were averaged, the RGB models performed the best (mAP of 95.1%), followed by the LWIR models (mAP of 94.5%), and finally the RGB-LWIR models (mAP of 92.6%). All test images in the testing-batch where daytime images. These results are not holistically accurate and indicative of how the models will perform in a real-world environment. Section C, D, and E of the results section discusses model performance against video footage.

| Overall results of ground-based and air-based ML model performance with image processing ||||||||
|---|---|---|---|---|---|---|---|
| Platform | Sensor | Object Class | mAP@0.5 | Precision (%) | Recall (%) | mAP@0.5 (%) | Averaged mAP |
| Ground Based | RGB | Car | 96.5% | 92.1% | 93.8% | 97.3% | 97.9% |
| | | Person | 97.0% | | | | |
| | | Truck | 98.5% | | | | |
| | LWIR | Car | 97.3% | 95.3% | 93.5% | 98.1% | |
| | | Person | 97.5% | | | | |
| | | Truck | 99.6% | | | | |
| | RGB+LWIR | Car | 96.6% | 95.1% | 97.2% | 98.4% | |
| | | Person | 99.0% | | | | |
| | | Truck | 99.6% | | | | |
| Air Based | RGB | Car | 91.3% | 83.9% | 91.2% | 92.9% | 90.2% |
| | | Person | 88.9% | | | | |
| | | Truck | 98.5% | | | | |
| | LWIR | Car | 88.9% | 82.7% | 85.1% | 90.9% | |
| | | Person | 85.1% | | | | |
| | | Truck | 98.8% | | | | |
| | RGB+LWIR | Car | 89.4% | 83.6% | 82.4% | 86.9% | |
| | | Person | 76.4% | | | | |
| | | Truck | 94.9% | | | | |

TABLE 2. Object class performance metrics with average precision, recall and mAP.

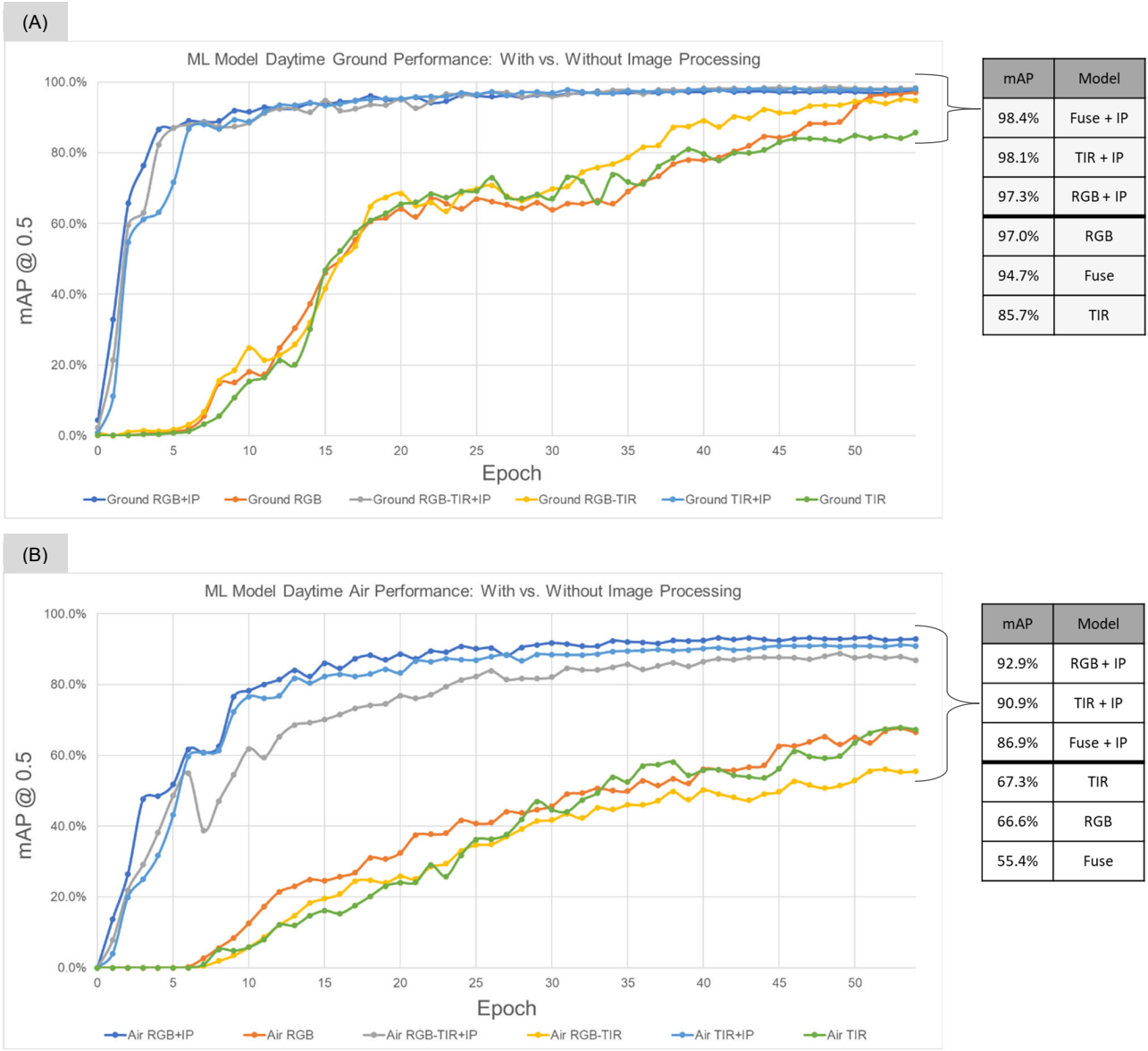

**FIGURE 7. Model training by epoch for all twelve models.**

*A. PERFORMANCE OF GROUND-BASED MODELING APPROACHES*

After all ground-based models underwent 55 epochs of training, the models with IP performed 5.4% better than the models with no IP. When integrating image processing into the fused RGB-LWIR approach (G_RGB-LWIR_IP) a high mAP of 98.4% was achieved. The LWIR model (G_LWIR_IP) performed similarly as the fused model with a mAP of 98.1%. The RGB approach (G_RGB_IP) performed well with a mAP of 97.3%.

For models without IP, the RGB model (G_RGB) had a mAP of 97%, while the fused model (G_RGB-LWIR) had a mAP of 94.7%. The LWIR approach (G_LWIR) performed the worst with a mAP of 85.7%.

When visualizing the mAP by epoch for each model during the training phase, a clear pattern emerges between the models with IP and the models without IP (figure 7). The models with IP all achieved a mAP of over 85% by the 11th epoch, while only two of the non-IP models surpassed the 85% mAP at the 49th epoch. This sharp increase of mAP in the models with IP in such a short number of epochs is mostly due to the larger quantity of images in the training dataset. The IP models had 2,100 total images for train-validate-training, while the non-IP models only had 300 original images for train-validate-training. The IP models were also trained on modified images with emphasized edges of object classes, thus enhancing ML object detection performance with many fewer epochs when compared to models trained without IP images.

Although the mAP for all ground-based models by the 55th epoch was within a 12.7% range, the three ML models with IP underwent a much more efficient training cycle than the three ML models without IP. The reason for this is likely due to the introduction of false positives within the non-IP models. Manually reviewing the model results on video footage revealed a high number of false positives in the non-IP approaches. Object detection models without IP images applied during training had a higher number of false positives, resulting in an overall lower mAP. This also resulted in more unstable precision and recall curves for the ML models without IP (figure 8). The ML models with IP had a significantly higher harmonic balance between precision and recall (F1 score) when compared to the models without IP (figure 8). Overall, the training process for models with IP was much less turbulent when compared to models with no IP.

### B. PERFORMANCE OF AIR-BASED MODELING APPROACHES

The contrast of mAP performance between models utilizing IP and those without was much greater in air-based then in ground-based approaches. In ground-based models there was a mere 5.4% difference between the average performance with IP versus without IP. In air-based models there was a 27.1% difference between the average performance with IP versus without.

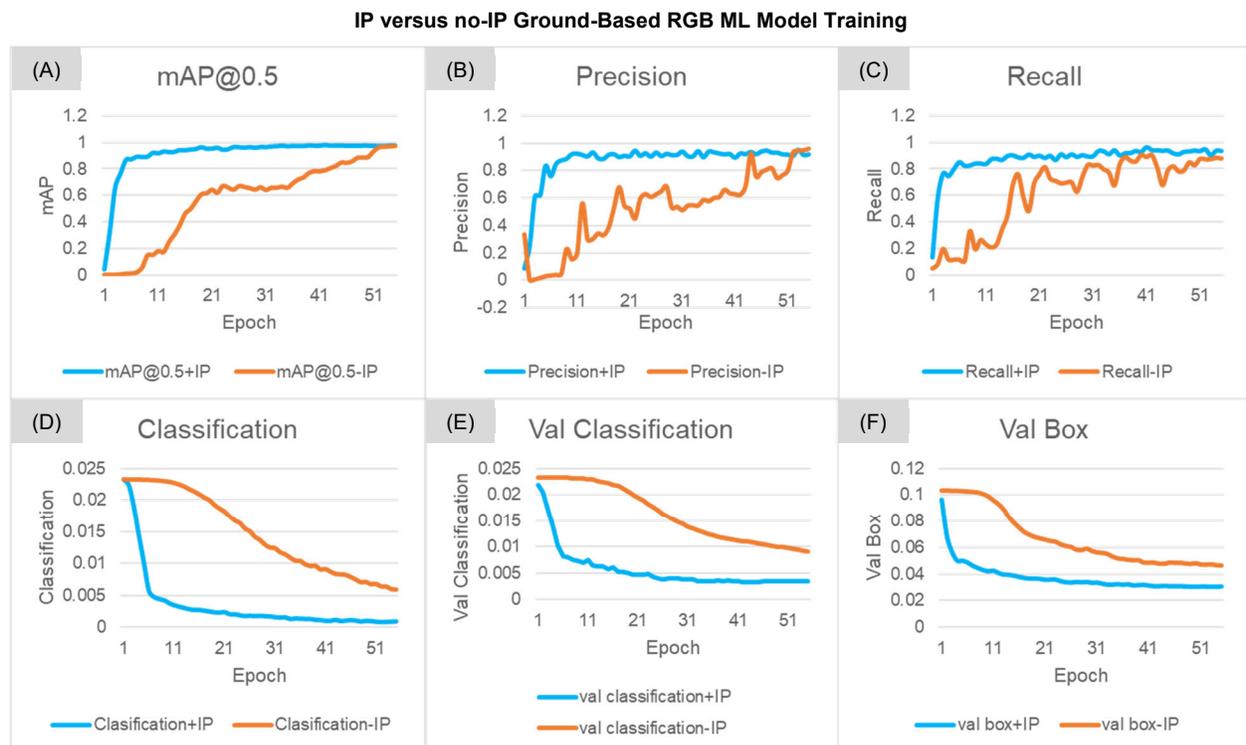

**FIGURE 8.** ML Training metrics comparing IP versus no IP.

The A_RGB_IP model performed the best with a mAP of 92.9%, followed by the A_LWIR_IP model with a mAP of 90.9%. The A_RGB-LWIR_IP model had a mAP of 86.9%. It is unknown why the performance gap was double between the A_RGB-LWIR_IP model and the A_LWIR_IP model (4% difference) when compared to the 2% performance gap between A_LWIR_IP and A_RGB_IP models. Air-based models with IP had an average mAP of 90.2%.

Non-IP air-based models performed significantly worse when compared to non-IP ground-based models. The A_LWIR model was the best performing non-IP air-based model with a mAP of 67.3%, followed by the A_RGB model with a mAP of 66.6%. The A_RGB-LWIR model performed the weakest with a mAP of 55.4%. In both IP and non-IP air-based ML models the A_RGB-LWIR models had the poorest performance.

### C. TESTING GROUND-BASED ML MODEL INFERENCE ON VIDEO

After the twelve ML models underwent training, validation and testing within the YOLOv7 notebook, model performance was tested against video footage that the model has never seen before. The six ground models were tested against video footage that contained all three object classes. The video was recorded in Dupont Circle in Washington D.C. in September 2022. The temperature was 28° C (82° F). A portion of the video was also darkened by shade from the trees, allowing LWIR cameras to demonstrate their added potential in situations where RGB cameras cannot see. All inference videos from this research can be viewed publicly [94]. A counting algorithm was also applied to the videos to help count total detections per frame. The counting algorithm counts all inferences to include false positives.

As indicative of the mAP results from YOLOv7 training, all six ground-based models performed very similar when conducting inference on video footage. The primary differences were detections in darker areas, detections at long distances, and confidence levels. In the darker areas of the video the G_RGB-LWIR_IP model performed the best at detecting humans (figure 9). The RGB models struggled to detect humans because of the lack of well-defined edges. Moreover, when detecting objects at long distances both IP and non-IP models for RGB, RGB-LWIR and LWIR performed almost equally, with the non-IP models sometimes performing better at detecting cars at longer distances. The G_RGB-LWIR model had the highest confidence levels of all six models. The RGB-LWIR model had approximately 1%-5% higher confidence scores then the RGB or LWIR models. The G_RGB-LWIR_IP model also had almost identical confidence scores of the G_LWIR_IP model when detecting humans in the darker areas of the video. The G_RGB-LWIR_IP model was also the best model at differentiating trucks from cars and had the highest confidence level when conducting inference on trucks.

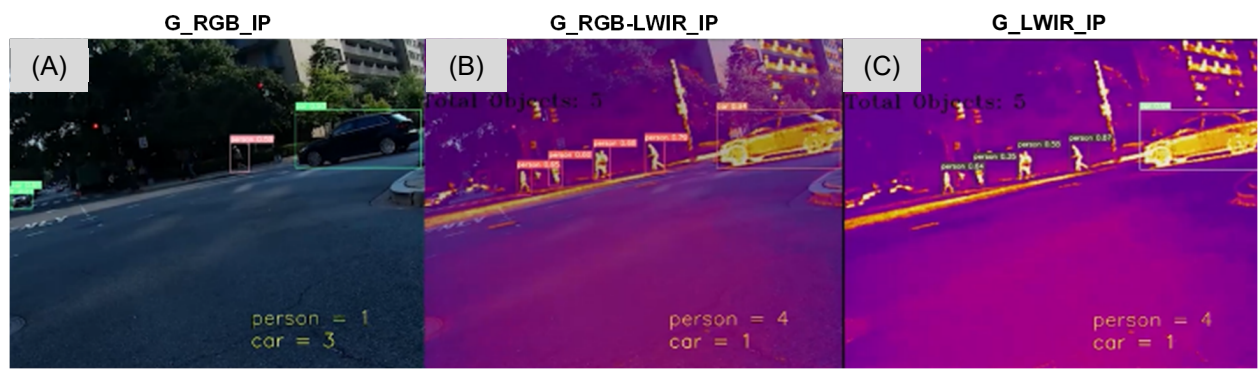

FIGURE 9. Ground-based ML models with IP conducting inference between RGB, RGB-LWIR, and LWIR models.

### D. TESTING AIR-BASED ML MODEL INFERENCE ON VIDEO

All six air-based models were evaluated against video footage of highway traffic in northern Virginia at noon. The temperature was 13° C (55° F) and the altitude of the multirotor drone was 45 m (150 ft). The model that performed the best was the A_RGB_IP which had on average a 20% higher confidence level then the A_RGB model. Although the A_RGB_IP and A_RGB models both had the same number of detections, the A_RGB_IP model had an abnormally high number of truck false positives. When a car would enter the frame the A_RGB_IP model would momentarily detect the car as a truck before correcting itself to car.

The A_RGB-LWIR models were the next best performing instances. A_RGB-LWIR_IP and A_RGB-LWIR models had equal performance with some differences. The A_RGB-LWIR_IP model was able to detect cars in the closer lane with 10%-15% higher confidence then the A_RGB-LWIR model. The A_RGB-LWIR model had increased car detections in a traffic lane that was farther away with 5%-10% higher confidence then the A_RGB-LWIR_IP model. The reason for this is unknown, but parallax most likely played an affect.

A_LWIR_IP and A_LWIR models performed the worst. In some frames the A_LWIR_IP model was detecting approximately half the number of cars when compared to the A_RGB_IP model. The confidence levels for the A_LWIR_IP model ranged between 35% to 90% while the confidence levels for the A_RGB_IP model ranged between 75% to 95%. The poor performance of the A_LWIR models was likely due to a combination of low thermal resolution and thermal crossover as a result of the warm highway asphalt.

A night test was then conducted against all six air-based models. The night footage utilized was the same highway traffic in northern Virginia that was used to evaluate the air-based day models. The footage was collected just after sunset, so thermal crossover from the highway was still high. The temperature was 7° C (45° F) and the altitude was 60 m (200 ft). When testing the model against night-time air footage, the A_LWIR_IP and A_RGB-LWIR_IP models performed the best under different circumstances. The A_LWIR_IP model had the highest detection rate when looking at oncoming traffic (front and back of cars). The A_LWIR_IP model detected approximately one-third of all the cars and had a confidence level between 30% and 90%. The A_LWIR_IP and A_RGB-LWIR_IP models also had significantly fewer false-positives when compared to the models without IP. When the multirotor was rotated 90° to collect video on a predominantly forested area adjacent to the highway, the A_LWIR and A_RGB-LWIR models had a very high number of false-positives in the forested areas of the video, detecting some of the trees as cars. The A_LWIR_IP and A_RGB-LWIR_IP models had no false positives in the woods adjacent to the highway.

When the multirotor drone was turned to observe a side-facing view of the highway to maximize the side-views of the vehicles, the A_RGB-LWIR_IP model performed the best, having almost five times the number of detections when compared to the A_LWIR_IP model. The reason for this is likely because there were more identifiable edges for ML models to detect cars. Why the A_RGB-LWIR_IP model was able to identify more vehicles from a right-angle view then the A_LWIR_IP model is unknown. A theory is that the darkened RGB footage in the A_RGB-LWIR_IP fusion provided an optimal blend that dampened the effects of thermal crossover. Under night-time conditions the A_RGB models unsurprisingly performed the worst. Both A_RGB models had very few car detections with a heavy amount of person false positives.

### *E. INFERENCE FROM VARIOUS ELEVATIONS*

The final multirotor test was to measure model performance at various elevations to measure performance decrease. The ML models were tested against footage recorded at 30 m (100 ft), 45 m (150 ft), 61 m (200 ft), 76 m (250 ft), 91 m (300 ft), 106 m (350 ft), and 121 m (400 ft). Footage could not be collected above 121 m due to Federal Aviation Administration (FAA) drone regulation that prohibits drones from flying above 121 m. The test area was a parking lot in front of a hardware store in Fairfax, Virginia, where all three object classes were present. The footage was recorded early in the morning with an ambient temperature of 2° C (36° F). This cold temperature combined with an early-morning collection period allowed for minimal thermal crossover effects. The cold weather also highlighted warm vehicles vs cold vehicles in LWIR imagery. Only models with IP were used for this evaluation.

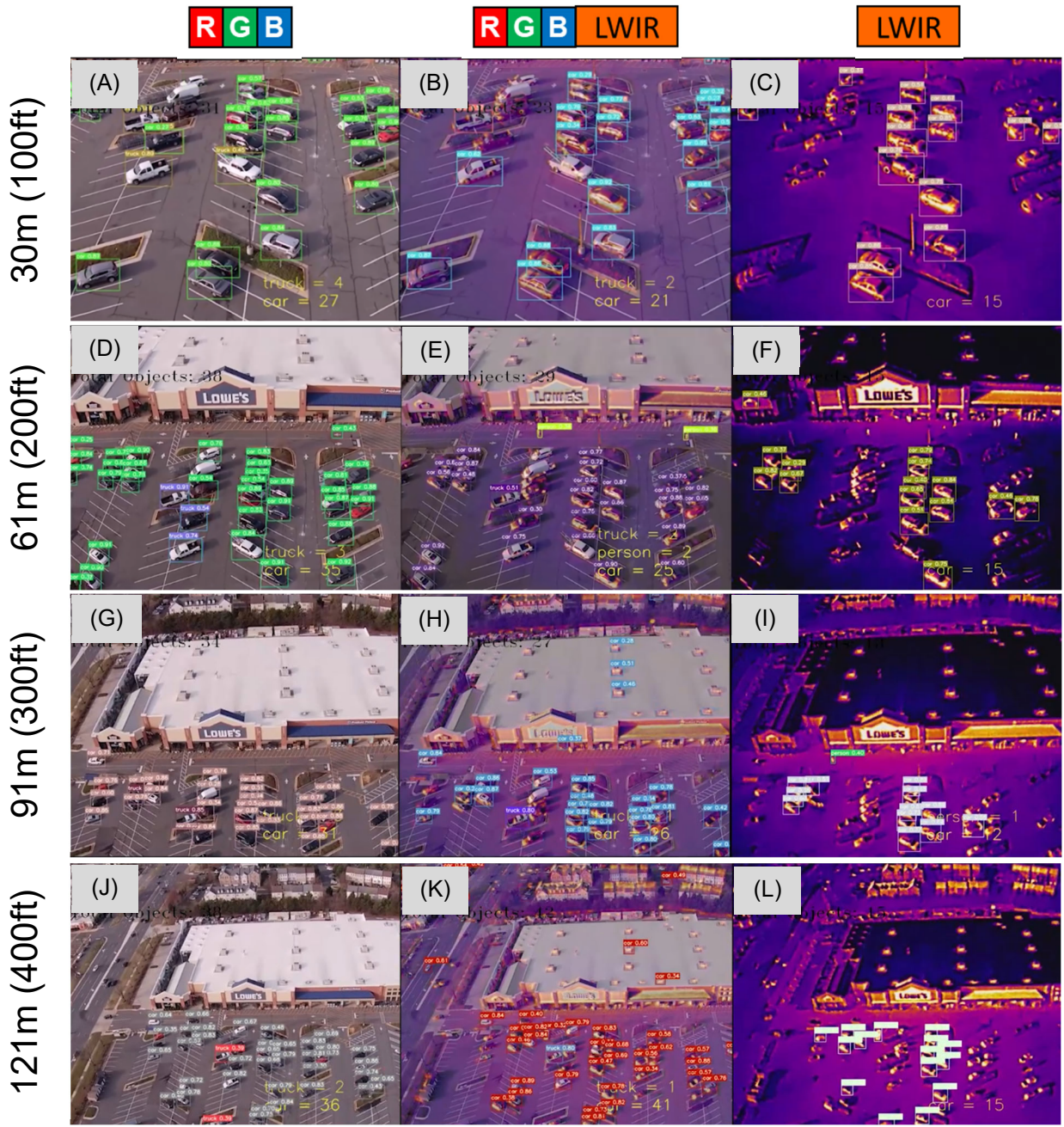

**FIGURE 10.** ML model performance at 30 m, 61 m, 91 m, and 121 m at a parking lot where all object classes were present.

At 30 m (100 ft) the A_RGB_IP model performed the best with the highest detection rates of both cars and trucks. All cars and trucks were accurately classified by the A_RGB_IP model. The A_RGB-LWIR_IP model classified almost all the vehicles but had difficulty in identifying the trucks. The confidence levels between A_RGB_IP and A_RGB-LWIR_IP were similar. In A_LWIR_IP footage the sides of the vehicles that were facing the sun were the most identifiable by the A_LWIR_IP model. The sun warmed the sides of the vehicles, allowing for more distinct and unique edges (figure 10). The cars that were forward-facing the sun appeared colder and had an inverse-thermal crossover effect where the cold parking lot and cold sides of the cars reduced any meaningful contrast needed for the model to detect object classes.

At 45 m (150 ft) the A_RGB_IP model continued to perform the best in terms of distinguishing cars and trucks while maintaining a high confidence level. However, the A_RGB-LWIR_IP and A_LWIR_IP model were the only two models that were able to detect humans in the parking lot. Although the A_RGB-LWIR_IP model had slightly lower confidence scores when compared to the A_RBG_IP model, it had the added benefit of detecting humans. The A_LWIR_IP model continued to identify most of the vehicles that were side-facing the sun as well as humans. Interestingly, the sun side-facing vehicles in the A_LWIR_IP model had a marginally higher confidence score then the confidence scores of sun side-facing vehicles in the A_RGB_IP model.

From 61 m (200 ft) all three models performed similarly to model performance at 45 m (150 ft). 61 m was also the optimal altitude for the A_RGB_IP model to yield the highest confidence scores. Car detection confidence scores were between 75% to 90% confidence. However, at 76 m (250 ft) the A_RGB-LWIR_IP model began to frequently identify the squared-shaped cooling systems on the roof of the hardware store as cars (figure 10). These false positives were occurring frequently in the same area of the roof. The A_RGB_IP and A_LWIR_IP models did not encounter this issue. The A_RGB-LWIR_IP and A_LWIR_IP models were both still detecting humans at 76 m.

At 91 m (300 ft) and 106 m (350 ft) the A_RGB_IP model was still detecting all cars and trucks with very high accuracy (60%-80%). The A_RGB-LWIR_IP model also had relatively high detection rates and confidence scores but was still identifying the cooling systems on the hardware store roof as cars. The A_LWIR_IP model continued to identify mostly sun side-facing vehicles.

Finally, at 121 m (400 ft) the A_RGB_IP model continued to perform with very high accuracy. Every car in the parking lot was being detected. However, the trucks became too small in resolution and where more often being counted as cars instead of trucks. In some frames the model identified the trucks as trucks. At 121 m an adjacent road to the parking lot became visible in the footage. However, the A_RGB_IP model only detected three moving cars on the busy road. The A_LWIR_IP model continued to detect the sun side-facing vehicles in the parking lot and did not detect any of the moving vehicles on the road. The A_RGB-LWIR_IP model continued to detect all the vehicles in the parking lot as well as the moving cars on the road. Although there were also a number of car false positives next to the road where there were no cars, the A_RGB-LWIR_IP model was the only model that was able to successfully identify moving cars on the road at 121 m. The reason for this is unknown.

## V. DISCUSSION

The research question that this study sought to answer the following questions:

*How do fused RGB-LWIR object detection models perform against separate RGB and LWIR approaches?*

In daytime conditions the RGB-LWIR model performed equal to RGB inference while maintaining the strengths of the LWIR detections in darker areas where RGB would fail. Despite Mixed results generated within the YOLOv7 testing (RGB-LWIR was the best performing ground-based model and worst performing air-based model) RGB-LWIR had the most detections of any model based on video footage observations. The RGB-LWIR model is also the best model at detecting moving objects.

In nighttime conditions the RGB-LWIR model had more detections then the LWIR approach by a factor of 2-3 more detections per frame. The RGB model did not work during nighttime conditions. The RGB-LWIR approach is superior because if its ability to (i) conduct better inference at night then LWIR (ii) maintain equal detection rates during daytime as RGB (iii) and detect more moving objects then either the RGB or LWIR approaches in both daytime and nighttime conditions.

*What is the impact of image processing (IP) on RGB, LWIR and fused RGB-LWIR model performance?*

Image processing had a positive effect on all models. IP had the greatest affect in increasing mAP in air-based models. Within YOLOv7 training and in video footage, IP was directly responsible for decreasing false positives. An analysis of the video footage across all models and platforms clearly demonstrates IP benefits in making models more resilient and reliable. IP increased ground-based mAP by 5.4% and air-based mAP by 27.1%. The stark difference between the ground and air IP performance increase is due to prominent object class edges. The IP versus non-IP difference was minimal in ground models because object classes have more defining edges and contrasting backgrounds, thus allowing non-IP models to conduct inference with fairly high mAP (figure 11). Object classes viewed from air-based platforms have fewer defining edges combined with thermal crossover challenges. These air-based challenges allowed IP to dramatically increased mAP and inference in air models.

*In what way do different drone altitudes affect model performance?*

The two most prominent model reactions to increasing altitude are an increase in the number of false positives (mostly in the RGB-LWIR model) and the loss of differentiation between cars and trucks at 106 m. When manually analyzing drone footage at various elevations it became apparent that air-based model characteristics and behaviors were consistent through all altitudes. For example, issues with LWIR inference (i.e. detecting mostly sun side-facing vehicles) were enduring from 15 m to 122 m. The RGB-LWIR model suffered isolated false positives from 61 m and 122 m. The RGB model had the fewest inference issues and was the only model capable of differentiating cars and trucks from 15 m to 106 m.

In this paper a methodical data-collection plan and experiment model were designed to collect data using ground and airborne systems to build and measure the performance of twelve distinct ML models using three different image types. This approach was used to quantify how twelve distinct ML models performed in controlled conditions. This research contributes to the literature in that it provides benchmark-metrics for how RGB, LWIR, and RGB-LWIR sensors perform using a YOLOv7 object detection model in both day and night conditions from two different platforms. This research also quantified object detection performance metrics at fixed altitudes. This research is also novel in that it quantifies performance differences between how ML models perform with image processing versus models without image processing.

The ML model performance metrics from this research provides an insight into capabilities and limitations, as well as lessons-learned, for deploying ground or air-based multispectral object detection models. The results demonstrate that not one specific object detection model type is best suited for all conditions, and that each ML model type has its own strengths and weaknesses for certain situations. This section will revisit the research question posed at the introduction of the paper.

This research successfully quantified the performance-difference between ground-based and air-based ML models. Air-based ML models with IP had a mAP of 90.2% while ground based ML models with IP had a mAP of 97.9%, resulting in a 7.7% performance difference between ground and air models. The performance gap between ground and air-based models is most likely due to background contrast, edge differences and distance. Ground-based ML models will generally have higher confidence scores because the horizon/sky provides a better background for contrasting the object class when compared to air-based models (figure 11).

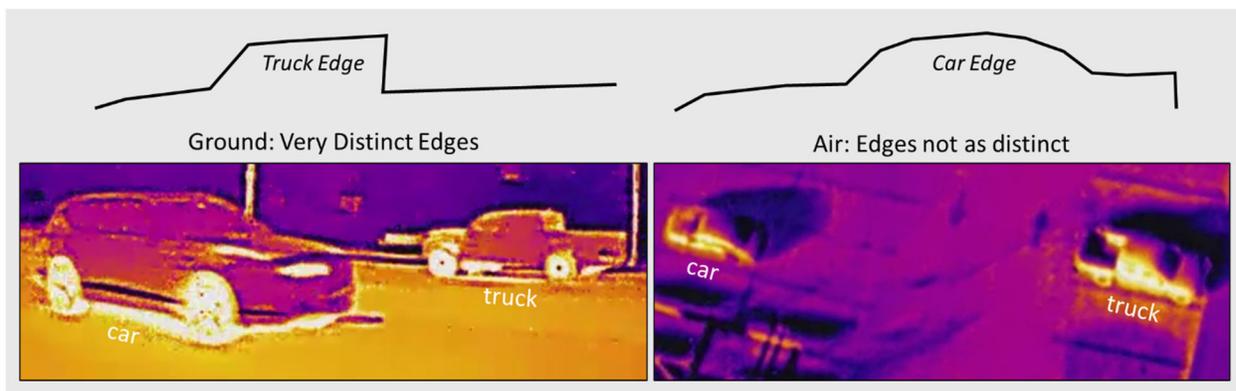

**FIGURE 11.** Comparing ground-based object-class edges versus air-based object class edges.

This holds especially true for LWIR and RGB-LWIR models when thermal crossover is a factor. Secondly, the object classes used for this research had much more distinct edges when looked at from the ground (figure 11). From the air it was apparent that the ML models struggled with differentiating between cars and trucks because of less-defined edges. The air-based ML models also conducted inference higher in altitude then the ground-based models conducted in horizontal distance.

This research also demonstrated the importance of applying image processing techniques when training models, especially when training air-based models. The six image processing techniques applied to the original images increased ground-based model mAP by 5.4% and air-based model mAP by a staggering 27.1%. The gap in IP performance between air and ground-based models is most likely due to air-based images having a less-contrasting background when compared to ground-based images (figure 11). Because of the less-contrasting background found in air-based images, IP had more contribution in increasing precision, recall, mAP, and many other metrics during training and deployment. IP was also able to bring both ground and air-based ML models to a very high mAP in shorter number of epochs. If more IP techniques were utilized the mAP would likely increase. A future area of study would be to train various ML models with different combinations of IP techniques and measure which IP techniques yield the highest mAP scores.

To answer the research question of determining which ML model performed the best, the results were not clear-cut and situation dependent. Although conducting a mAP evaluation of each model after training within the YOLOv7 python script is quantitatively the fairest way to evaluate and rank the models, testing the models against new footage to imitate performance in a deployed setting will provide much more relevant feedback on how the models behave in various environments and altitudes. Although all the ground-based models performed similarly within the YOLOv7 testing process, when deployed against new footage all models have their own pros and cons. For example, when planning to conduct ground-based collection in a place where there is a plethora of shadows, such as under a jungle canopy or in a city as the sun is setting or rising, the RGB-LWIR or LWIR models should be selected. Conversely if collection is conducted in an open area with minimal shadows and high temperatures then the RGB model should be used. This research did demonstrate that the RGB-LWIR models are the most resilient model out of the three models due to its ability to work in all conditions no matter the illumination. Although the RGB-LWIR did suffer minor false-positive issues, if one model had to be selected to improve upon for future research and advancing the study of object detection then then RGB-LWIR model is the logical choice.

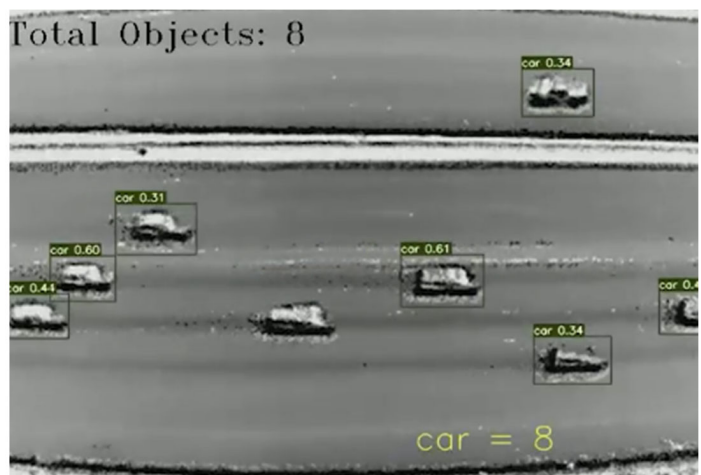

**FIGURE 12.** LWIR ML model trained on fusion palette conducting inference on video using the black-hot palette.

Another benefit of the dataset generated from this research is that the thermal imagery collected is cross-compatible with other thermal palettes. White-hot and black-hot thermal palettes are the most commonly utilized palettes. These are the LWIR palettes that the U.S. military use. These palettes also provide excellent contrast. White-hot and black-hot were not used for this research because they did not create a good contrast with RGB imagery. The LWIR and RGB-LWIR labeled dataset with the fusion palette can be used to detect thermal objects in other LWIR pallets such as white-hot and black-hot palettes. A test was conducted to confirm cross-compatibility, and the confidence levels were almost identical as the fusion-based models (figure 12). Future research can be conducted into measuring ML models against different thermal palettes, as well as combining different fusion levels to determine the optimal RGB-LWIR fusion level for a specific time of day.

## VI. CONCLUSION

In this study twelve unique object detection YOLOv7 models were built to conduct inference from ground and air platforms. Six IP techniques were also applied to six models. IP inference performance was measured and observed against the models without IP to demonstrate the unique advantages of applying edge-enhanced images to object detection applications. This study established that six IP techniques increased ground-based mAP by 5.4% and air-based mAP by 27.1%. This research also codified model metrics for RGB, LWIR and RGB-LWIR object detection applications. Baseline performance metrics were measured for all three applications, resulting in a RGB mean average precision (mAP) of 95.1%, a LWIR mAP of 94.5%, and a blended RGB-LWIR mAP of 92.6%. When conducting real-time inference, the RGB-LWIR model is assessed as the superior approach because of its ability to conduct better inference at night, maintain equal detection rates during daytime as RGB, and increased inference detection in moving objects in both day and night conditions. Above all, this research cultivated a 12,600 labeled image dataset for ground-based and air-based RGB, LWIR, and RGB-LWIR fused imagery to encourage further multispectral machine-driven object detection research from both air and ground platforms.

The rapid growth and synthesis of both UAS and object detection technologies will continue to make the study of UAS-borne ML model performance of vital importance. Motivated by this development, this research applied a framework to collect data using ground and airborne systems to build and measure the performance of twelve distinct models using three different image types. This approach is the first of its kind in that all three image types were collected, extracted, processed, and labeled in the exact same manner in an attempt to limit external variables from convoluting the results. This approach also did not stop testing the ML models at the testing phase of the YOLOv7 testing batch, but rather continued to apply the ML models to new video to observe performance over various scenes and altitudes. The results from this research provide a quantitative bedrock to help springboard future research into LWIR-related object detection from both ground and air systems.

The results of the research revealed that applying six unique image processing techniques to ML model training increased ground-based model performance by 5.4% and air-based model performance by 27.1%. The research also quantified a 7.7.% loss in performance when transitioning a model from a ground-based to an air-based platform. Lastly, this research showed that the blended RGB-LWIR model is the most capable and resilient model when deployed in dynamic environments where temperature and illumination factors are everchanging or unknown. However, the RGB-LWIR model is not always the most accurate model to select for the task. The results of this research will help guide the decision-making cycle into determining what kind of object detection model should be deployed from a ground-based or air-based platform to maximize detections. Indeed, the integration of LWIR imagery will enhance both human and machine three-dimensional (3D) depth perception, compared to traditional RGB-imagery methods, providing an overall increase in situational awareness.

## ACKNOWLEDGMENT

The authors would like to thank the Geography & Geoinformation Science Department at George Mason University for supporting and funding this research. JEG would like to thank the United States Army for resource contributions that allowed him to carry out this research.

## COMPETING INTERESTS AND AUTHOR CONTRIBUTION

The authors declare no competing interests. JEG focused on data collection, processing, model creation and paper writing, while EJO focus was on providing research direction, project management and writing the final manuscript.

## APPENDIX A: DATA AVAILABILITY

All data and models can be found in the below Zenodo links.

[Air-based labeled data for all object classes](#)
[Ground-based labeled data for all object classes](#)

[Air and ground-based ML Model weights](#)
[Image Processing Code](#)
[YOLOv7 Training Code](#)
[Inference Videos](#)

## APPENDIX B: REVIEWED PAPERS

Table A.1: Key information from reviewed papers by data type.

Categories: RGB, LWIR, Object Detection (OD), UAS, Machine Learning (ML)
Sensor type: RGB, LWIR, RGB-LWIR

| Data Type | Topic | Publication | Category | Sensor type |
|---|---|---|---|---|
| Journal Article | OD with RGB/LWIR | Tu et al., 2021 | RGB, LWIR, OD | RGB/LWIR |
| Conference Paper | LWIR mapping | Vidas et al., 2013 | RGB, LWIR | RGB, LWIR |
| Journal Article | RGB, LWIR and OD | Teju and Bhavana, 2020 | RGB, LWIR, OD | RGB, LWIR |
| Conference Paper | OD with RGB/LWIR | St-lauren et al., 2007 | RGB, LWIR, OD | RGB, LWIR |
| Journal Article | OD with LWIR on UAS | Li et al., 2021 | RGB, LWIR, OD, UAS | RGB, LWIR |
| Journal Article | RGB-LWIR fusion | Alldieck et al., 2016 | RGB, LWIR | RGB, LWIR |
| Journal Article | OD with RGB/LWIR | Lia et al., 2022 | RGB, LWIR, OD | RGB, LWIR |
| Journal Article | OD with RGB/LWIR on UAS | Speth et al., 2022 | RGB, LWIR, OD, UAS | RGB, LWIR |
| Journal Article | OD with RGB/LWIR | Chen et al., 2022 | RGB, LWIR, OD | RGB, LWIR |
| Journal Article | OD with RGB/LWIR on UAS | Yang et al., 2022 | RGB, LWIR, OD, UAS | RGB, LWIR |
| Journal Article | RGB/LWIR and edge detection | Zhou et al., 2021 | RGB, LWIR, OD | RGB, LWIR |
| Conference Paper | RGB/LWIR semantic segmentation | Deng et al., 2021 | RGB, LWIR, OD | RGB, LWIR |
| Journal Article | RGB/LWIR semantic segmentation | Sun et al., 2021 | RGB, LWIR, OD | RGB, LWIR |
| Journal Article | RGB-LWIR fusion | Fendri et al., 2017 | RGB, LWIR, OD | RGB, LWIR |
| Journal Article | RGB/LWIR semantic segmentation | Zhou et al., 2021 | RGB, LWIR, OD | RGB, LWIR |
| Journal Article | OD with RGB/LWIR | Zhou et al., 2022 | RGB, LWIR, OD | RGB, LWIR |
| Conference Paper | OD with RGB/LWIR | Bañuls Arias et al., 2020 | RGB, LWIR, OD | RGB, LWIR |
| Journal Article | OD with RGB/LWIR | Luo et al., 2014 | RGB, LWIR, OD | RGB, LWIR |
| Conference Paper | OD with RGB/LWIR to detect oil leaks | Lie at al., 2019 | RGB, LWIR, OD | RGB, LWIR |
| Journal Article | OD with RGB/LWIR | Guo et al., 2021 | RGB, LWIR, OD | RGB, LWIR |
| Journal Article | OD with RGB/LWIR | Sun et al., 2019 | RGB, LWIR, OD | RGB, LWIR |
| Journal Article | OD with RGB/LWIR | Krišto et al., 2020 | RGB, LWIR, OD | RGB, LWIR |
| Journal Article | RGB/LWIR OD from UAS | Fei et al., 2022 | RGB, LWIR, OD, UAS | RGB, LWIR |
| Journal Article | RGB/LWIR OD from UAS | De Oliveira and Wehrmeister, 2018 | RGB, LWIR, OD, UAS | RGB, LWIR |
| Journal Article | OD with RGB/LWIR for tree diameter | da Silva et al., 2021 | RGB, LWIR, OD | RGB, LWIR |
| Conference Paper | OD with RGB/LWIR | Lahuad and Ghanem, 2017 | OD | RGB |
| Journal Article | OD with UAS | Tian et al., 2021 | OD, UAS | RGB |
| Journal Article | OD with UAS | Akbari et al., 2021 | OD, UAS | RGB |
| Conference Paper | UAS camera stabilization | Rajesh and Kavitha, 2015 | UAS | RGB |
| Journal Article | OD with RGB | Roy, 2022 | RGB, OD | RGB |

| Type | Topic | Authors | Keywords | Category |
|---|---|---|---|---|
| Journal Article | OD with RGB on UAS | Kazaz et al., 2021 | RGB, OD, UAS | RGB |
| Journal Article | OD with RGB on UAS | Hossain and Lee | RGB, OD, UAS | RGB |
| Journal Article | RGB OD from UAS | Xiaoliang Wang at al., 2019 | RGB, OD, UAS | RGB |
| Journal Article | RGB OD from UAS | Ezzy et al., 2021 | RGB, OD, UAS | RGB |
| Journal Article | RGB OD from UAS | Hyun-Ki Jung and Gi-Sang Choi, 2022 | RGB, OD, UAS | RGB |
| Journal Article | OD with RGB | Huang et al., 2022 | RGB, OD | RGB |
| Journal Article | OD on UAS with transfer learning | Walambe et al., 2021 | RGB, OD, UAS | RGB |
| Journal Article | UAS applications | Morales et al., 2019 | UAS | RGB |
| Conference Paper | OD with RGB | Liu et al., 2018 | RGB, OD | RGB |
| Journal Article | OD with RGB and edge detection | Ushma, 2022 | RGB, OD | RGB |
| Conference Paper | OD with RGB | Galvez et al., 2018 | RGB, OD | RGB |
| Journal Article | OD with RGB | Diwan et al., 2022 | RGB, OD | RGB |
| Journal Article | OD with RGB | Zhou et al., 2015 | RGB, OD | RGB |
| Conference Paper | OD with RGB | Lie et al., 2021 | RGB, OD | RGB |
| Conference Paper | OD with UAS flight control | Meier at al., 2011 | OD, UAS | RGB |
| Conference Paper | RGB stereoscopic images | Bergerson, 2010 | RGB | RGB |
| Journal Article | OD with RGB and blur | Wu et al., 2020 | RGB, OD | RGB |
| Conference Paper | OD with RGB from UAS | Koskowich et al., 2018 | RGB, OD, UAS | RGB |
| Journal Article | OD with Scikit-learn | Hao and Ho, 2019 | OD | RGB |
| Conference Paper | OD in military | Ozbay and Sahingil, 2017 | LWIR, OD | LWIR |
| Conference Paper | LWIR obstacle regognition | Cho, 2019 | LWIR | LWIR |
| Conference Paper | OD with LWIR | Sachan et al., 2022 | LWIR, OD | LWIR |
| Journal Article | LWIR deblurring | Rai et al., 2022 | LWIR | LWIR |
| Conference Paper | OD with LWIR | Nguyen and Tran, 2015 | LWIR, OD | LWIR |
| Journal Article | Athmosphrics and LWIR | Uzun et al., 2022 | LWIR | LWIR |
| Conference Paper | OD and LWIR | Nirgudkar and Robinette, 2021 | LWIR, OD | LWIR |
| Conference Paper | LWIR palette selection | Agrawal and Karar, 2018 | LWIR | LWIR |
| Journal Article | OD with LWIR | Pavlović et al., 2022 | LWIR, OD | LWIR |
| Journal Article | LWIR deblurring | Batchuluun et al., 2021 | LWIR, OD | LWIR |
| Conference Paper | OD and LWIR | Khatri et al., 2022 | LWIR, OD | LWIR |
| Journal Article | LWIR and UAS | Driggers et al., 2021 | LWIR, UAS | LWIR |
| Conference Paper | LWIR OD from UAS | Juang et al., 2020 | LWIR, OD, UAS | LWIR |
| Conference Paper | LWIR and edge detection | Sergyán, 2012 | LWIR, OD | LWIR |
| Journal Article | enhancing OD with LWIR | Agrawal and Subramanian, 2019 | LWIR, OD | LWIR |
| Journal Article | OD with LWIR | Choi et al., 2018 | LWIR, OD | LWIR |
| Journal Article | Countermeasures to LWIR detection | Shen et al., 2015 | LWIR | LWIR |
| Conference Paper | OD with LWIR | Nan et al., 2018 | LWIR, OD | LWIR |
| Conference Paper | OD with LWIR | Korobchynskyi et al., 2018 | LWIR, OD | LWIR |
| Conference Paper | OD with LWIR | El Ahmar et al., 2022 | LWIR, OD | LWIR |
| Thesis | OD with LWIR | Bhusal, 2022 | LWIR, OD | LWIR |

| Type | Topic | Author | Tags | Category |
|---|---|---|---|---|
| Journal Article | LWIR OD from UAS | Jiang et al., 2022 | LWIR, OD, UAS | LWIR |
| Journal Article | LWIR OD from UAS | Leira et al., 2021 | LWIR, OD, UAS | LWIR |
| Conference Paper | OD with LWIR and Saliency | Altay and Velipasalar, 2020 | LWIR, OD | LWIR |
| Conference Paper | OD model comparisson with LWIR | Sharrab et al., 2021 | LWIR, OD | LWIR |
| Conference Paper | OD with LWIR | Wei, 2021 | LWIR, OD | LWIR |
| Conference Paper | LWIR semantic segmentation | Shivakumar et al., 2020 | LWIR, OD | LWIR |
| Conference Paper | Synthetic LWIR generation | Blythman et al., 2020 | LWIR, OD | LWIR |
| Journal Article | LWIR and thermal crossover | Zhao et al., 2016 | LWIR | LWIR |
| Journal Article | LWIR and surveys | Gade and Moeslund, 2014 | LWIR | LWIR |
| Journal Article | Countermeasures to LWIR detection | Qu et al., 2018 | LWIR | LWIR |
| Conference Paper | OD with LWIR | Setjo et al., 2017 | LWIR, OD | LWIR |
| Journal Article | OD with LWIR for autonomous driving | Dai et al., 2021 | LWIR, OD | LWIR |
| Journal Article | LWIR image generation for OD | Lie et al., 2021 | LWIR, OD | LWIR |
| Conference Paper | UAS proliferation | Wargo et al., 2016 | UAS | |
| Conference Paper | UAS shortcomings | Wargo et al., 2018 | UAS | |
| Journal Article | UAS outlook | Canis, 2022 | UAS | |
| Journal Article | OD training fundamentals | Brownlee | OD | |
| Journal Article | OD YOLOv7 | Wang et al., 2022 | OD | |
| Book | ML planning | Oughton et al., 2016 | ML | |
| Journal Article | ML planning | Aslani et al., 2022 | ML | |
| Journal Article | ML planning | Oughton and Mathur, 2021 | ML | |

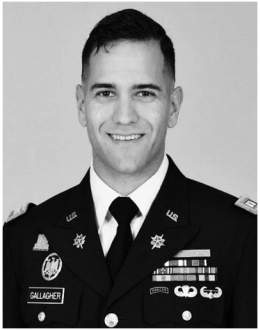

**JAMES E. GALLAGHER** received a B.A from Mercyhurst University in Intelligence Studies and a M.A from American Military University in Intelligence Studies. In 2021 he began a M.S. in Geoinformatics and Geospatial Intelligence Studies at George Mason University. He is also an active-duty U.S. Army Captain in the Military Intelligence Corps. CPT Gallagher's research in RGB-LWIR object detection was inspired from his lessons-learned in Kirkuk Province, Iraq, where he used a variety of military sensors to locate Islamic State insurgents.

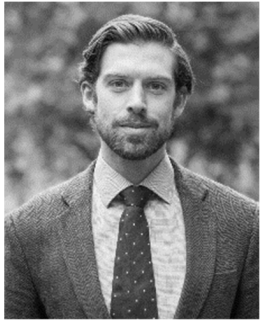

**EDWARD J. OUGHTON** received the M.Phil. and Ph.D. degrees from Clare College, at the University of Cambridge, U.K., in 2010 and 2015, respectively. He later held research positions at both Cambridge and Oxford. He is currently an Assistant Professor in the College of Science at George Mason University, Fairfax, VA, USA, developing open-source research software to analyze digital infrastructure deployment strategies. He received the Pacific Telecommunication Council Young Scholars Award in 2019, Best Paper Award 2019 from the Society of Risk Analysis, and the TPRC48 Charles Benton Early Career Scholar Award 2021.